\documentclass[10pt,twocolumn,letterpaper]{article}

\usepackage[pagenumbers]{cvpr} 

\usepackage[percent]{overpic}   
\usepackage{subcaption}         

\definecolor{cvprblue}{rgb}{0.21,0.49,0.74}
\usepackage[pagebackref,breaklinks,colorlinks,allcolors=cvprblue]{hyperref}

\def\paperID{22252} 
\def\confName{CVPR}
\def\confYear{2026}

\title{Arcee: Differentiable Recurrent State Chain for Generative Vision Modeling with Mamba SSMs}

\author{Jitesh Chavan$^{1}$ \ Rohit Lal$^{2,*}$ \ Anand Kamat$^{3,*}$ \ Mengjia Xu$^{1}$ \\\
$^{1}$New Jersey Institute of Technology$ \quad \quad ^{2}$University of California, Riverside $\quad \quad ^{3}$McGill University \\
{\tt\small jsc78@njit.edu, rlal011@ucr.edu, anand.kamat@mail.mcgill.ca, mengjia.xu@njit.edu} \\
{\small $^{*}$Equal contribution}
}
\begin{document}
\maketitle

\newcommand\blfootnote[1]{%
  \begingroup
  \renewcommand\thefootnote{}\footnote{#1}%
  \addtocounter{footnote}{-1}%
  \endgroup
}

\begin{abstract}
State-space models (SSMs), Mamba in particular, are increasingly adopted for long-context sequence modeling, providing linear-time aggregation via an input-dependent, causal selective-scan operation.
Along this line, recent “Mamba-for-vision” variants largely explore multiple scan orders to relax strict causality for non-sequential signals (e.g., images).
Rather than preserving cross-block memory, the conventional formulation of the selective-scan operation in Mamba reinitializes each block's state-space dynamics from zero, discarding the terminal state-space representation (SSR) from the previous block.
Arcee, a cross-block recurrent state chain, reuses each block's terminal state-space representation as the initial condition for the next block $(h_{0}^{(l)} = \mathcal{T}^{(l)}(h_{T}^{(l-1)}))$.
Handoff across blocks is constructed as a differentiable boundary map whose Jacobian enables end-to-end gradient flow across terminal boundaries.
Key to practicality, Arcee is compatible with all prior “vision-mamba” variants, parameter-free, and incurs constant, negligible cost.
As a modeling perspective, we view terminal SSR as a mild directional prior induced by a causal pass over the input, rather than an estimator of the non-sequential signal itself.
To quantify the impact, for unconditional generation on CelebA-HQ (256$\times$256) with Flow Matching, Arcee reduces FID$\downarrow$ from $82.81$ to $15.33$ ($5.4\times$ lower) on a single scan-order Zigzag Mamba baseline.
Extensible CUDA kernels and training code are released to support reproducibility and further research at \href{https://github.com/JiteshChavan/rc2}{https://github.com/JiteshChavan/rc2}

\end{abstract}

\section{Introduction}
\label{sec:intro}
Flow matching and diffusion models have revolutionized generative frameworks for images, videos, protein structures, and many other modalities (Lipman \etal, 2022 \cite{Lipman2022FM}; Albergo \etal 2023 \cite{Albergo2023SI}; Liu \etal \cite{Liu2022RectifiedFlow}; Bose \etal, 2024 \cite{Bose2024FoldFlow}; Song \etal \cite{Song2020ScoreSDE} 2020; Karras \etal \cite{Karras2022EDM}). These models generate realistic images, videos or samples from corresponding data distribution by simulating an ordinary or stochastic differential equation (ODE/SDE) with a sample from a simple prior (usually gaussian noise) as the initial value condition for the differential equation, where the vector field (and score function in case of SDEs) that defines the differential equation is approximated by a neural network. Recently, transformer architectures have proliferated as a choice for the neural network, a consequence of their superior scalability \cite{PeeblesDiT, Bao2023} and effectiveness in multi-modal training \cite{pmlr-v202-bao23a}. Despite their effectiveness for in-context learning tasks in non sequential modalities, transformers bear a significant computational cost that scales quadratically with input sequence length. While there have been efforts to alleviate the quadratic complexity of the attention mechanism by instrumenting methods such as FlashAttention, FlashAttention 2 \cite{DaoFlashAttention2022, dao2024flashattention2}, it still remains the bottleneck for employing transformer-based models \cite{Vaswani2017Attention}.

State-Space Models have emerged as competitive architectures for long context sequence modeling, offering linear time information aggregation across input signals via continuous time State-Space transitions \cite{Gu2022S4D, Gupta2022DSS, Gu2022S4LongSequencesWithS4}. Recent work improves SSM robustness and efficiency through better initializations \cite{Gu2022HowToTrainYourHiPPO}, parametrizations \cite{Gu2022S4D}, diagonalizations \cite{Gupta2022DSS}, and recurrence parallelizations \cite{Gu2021LSSL}. \emph{Mamba} \cite{GuDao2023Mamba} in particular extends prior works, making SSMs more expressive with input dependent state-space transitions through hardware-aware and work-efficient selective scan, yielding linear scaling in sequence length. While selective scan mechanism introduced in Mamba excels at efficient long sequence modeling, its causal aggregation of information creates friction when adapting Mamba to non sequential modalities, motivating architectures that preserve efficiency while relaxing the inherent strict causality. Prior vision-SSM work typically flattens 2D signals into a token sequence $u$ and applies multiple scan orders within the same block, followed by simple feature fusion, adding parameters for each scan order \cite{Liu2024VMambaNeurIPS,visionMamba,Liu2024SwinUMamba,Ma2024UMamba}. A complementary strategy amortizes layerwise heterogeneous scan orders across depth with no per-block parameter increase, as in \emph{Zigma} \cite{ZigMaECCV}, where each causal scan manifold $u \mapsto y$, captures dependencies between tokens at varying degrees of spatial vicinity across layers.

Mamba was originally introduced for \emph{autoregressive} sequence modeling; consequently, most vision variants inherit a design in which each block’s state–space dynamics are initialized with a \emph{zero} state instead of retrieving global causal summary encoded within terminal state-space representation $h_{T}$ from previous block to avoid information leakage and preserve causality; This is restrictive for non-sequential signals (e.g., images), because $h_T$ summarizes a full pass over the input in a given scan order and thus encodes potentially useful global context that is currently discarded.

\begin{figure*}[t]
  \centering
  \begin{subfigure}[b]{0.48\textwidth}\phantomsubcaption\label{fig:ssm-a}\end{subfigure}
  \hfill
  \begin{subfigure}[b]{0.48\textwidth}\phantomsubcaption\label{fig:ssm-b}\end{subfigure}

  \includegraphics[width=\textwidth]{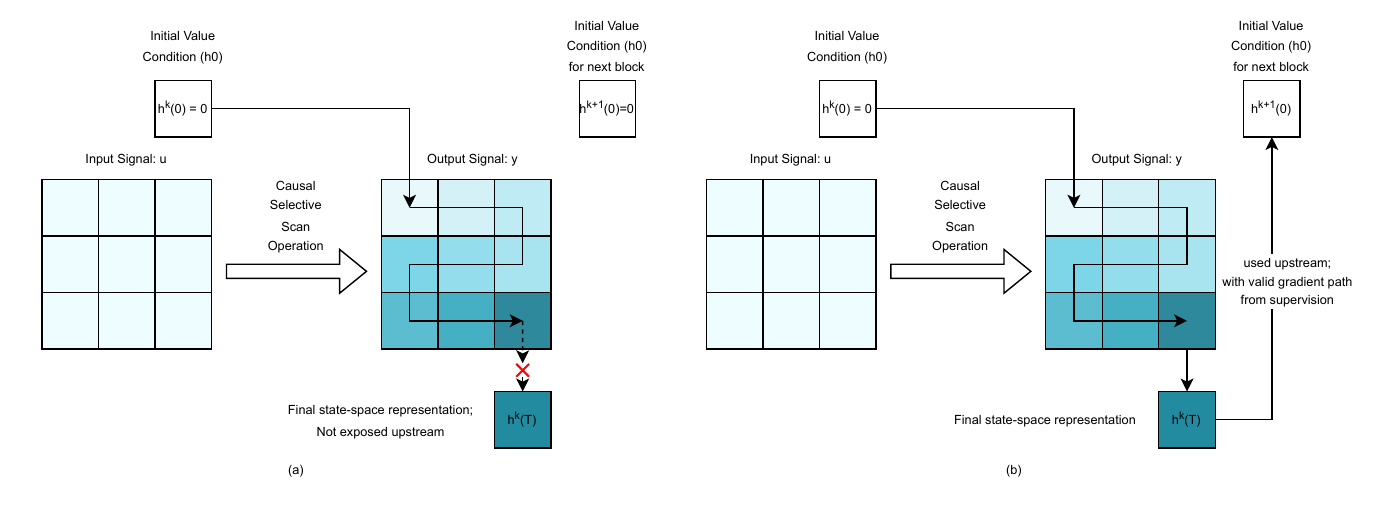}

  \caption{Conventional vs.\ Arcee selective scan. 
(a) In a vanilla Mamba block, the selective scan is strictly causal: the state is initialized with $h^{(k)}(0)=0$, the terminal state $h^{(k)}(T)$ is discarded after producing $y$, and the next block again starts from zero. Darker cells indicate positions that have accumulated more context (later timesteps have seen a larger prefix of the sequence). 
(b) Arcee extends the scan to a two-port block: the terminal SSR $h^{(k)}(T)$ is reused as the initial state $h^{(k+1)}(0)$ of the next block via a differentiable boundary map, creating a recurrent state chain across depth with a valid gradient path and no change to the intra-block dynamics.}

  \label{fig:selectivescanmanifold}
\end{figure*}

In this work, we introduce a cross-block \textbf{R}ecurrent State \textbf{C}hain (\textbf{Arcee}) that generalizes the conventional causal selective-scan in Mamba by using the terminal state-space representation (SSR) from block $l-1$ as the initial value condition for SSM dynamics in block $l$. Any prior Mamba baselines can be adapted to propagate this compact global summary across depth, yielding a plug-and-play mechanism with zero additional parameters and constant, negligible overhead (independent of sequence length). Arcee is orthogonal to scan-order design; whereas standard selective-scan fixes each block's initial state to \textbf{0}, Arcee reuses the previous block's final SSR to provide a mild, architecture-agnostic inductive bias for generative visions tasks with Mamba SSMs. To summarize, we make following contributions:
\begin{enumerate}
    \item We identify the legacy zero-init initial value condition for state-space dynamics in conventional selective scan operation in Mamba that discards the terminal state-space representation (SSR) between blocks (see \ref{fig:selectivescanmanifold}, restrictive for non-sequential signals (images).
    \item We hypothesize that the terminal SSR potentially encodes a useful global summary and serves as a mild inductive cue for downstream selective-scan dynamics, despite SSR being a severe compression of the non-sequential signal.
    \item We introduce a zero parameter overhead, plug and play solution \emph{Arcee} by generalizing the selective scan manifold in Mamba \cite{GuDao2023Mamba} from $u \mapsto y$ to $(u, h_{0}^{(l)}) \mapsto (y, h_{T}^{(l)})$ via a differentiable boundary map $h_{0}^{(l)} = \mathcal{T}^{(l)}(h_{T}^{(l-1)})$; its Jacobian $\mathcal{J}_{\mathcal{T}}^{(l)} = \partial h_{0}^{(l)}/\partial h_{T}^{(l-1)}$ is trained end to end so terminal SSRs $(h_{T}^{(k)}  \forall k \in [0, depth) )$ rendered by each block align across depth. In our default, $\mathcal{T}^{(l)}$ is identity.
    \item Plugging Arcee into naive single-scan order ZigZag-Mamba baseline (Zigma\_1) \cite{ZigMaECCV} for unconditional generation on CelebA-HQ (256x256) \cite{Karras2018PGGAN} with Flow Matching \cite{Lipman2022FM}, we observe a large FID drop from \textbf{82.81} to \textbf{15.33} (-81.49\%, \textbf{5.40x lower}) at negligible cost. As scan-order diversity increases (or in case of order-specific weights), the improvements attenuate; we analyze these trends in the experiments.
    \item We release a general Arcee-selective-scan CUDA implementation with exposed boundary hooks $(h_0, h_T)$ to enable research on cross-block state handoffs.
\end{enumerate}

\section{Background and Motivation}
\subsection{Generative Framework: Flow Matching}
\label{sec:FM}
This work employs the Flow Matching framework (Lipman \etal 2022 \cite{Lipman2022FM}; Albergo \etal 2023 \cite{Albergo2023SI}; Liu \etal \cite{Liu2022RectifiedFlow}) which learns a time–dependent vector field that transports a simple prior distribution $P_{\text{init}}$ (e.g., $\mathcal{N}\big(0,I_d\big)$) to the data distribution $p_{\text{data}}$ along the marginal probability path $P_{t}(x)$ by simulating an ODE.

We represent data points as vectors $z \in \mathbb{R}^d$ and the data distribution as $p_{\text{data}}$. The \textbf{Gaussian conditional probability path} $P_{t}(x_t \mid z)$ is constructed to interpolate between $\mathcal{N}(0, I_{d})$ and $\delta_{z}(\cdot)$ on interval $t \in [0, 1]$:
\begin{multline}
x_t = \alpha_t z + \sigma_t \epsilon, \quad \epsilon\sim\mathcal{N}(0,I_d) \\
\iff\quad p_t(x_t\mid z) = \mathcal{N}\big(x_t;\alpha_t z,\sigma_t^2 I_d\big)
\label{eq:ConditionalProbPath}
\end{multline}
where $\alpha_{t}, \sigma_{t} \geq 0$ are \textbf{interpolation schedulers} with $\alpha_{0} = \sigma_{1}=0$ and $\alpha_{1} = \sigma_{0}=1$ and $\alpha_t$ (resp. $\sigma_t$) strictly monotonically increasing (resp. decreasing). The conditional vector field for Gaussian conditional probability paths is given by (see \cite{Lipman2022FM, Lipman2024FMGuide}):
\begin{equation}
    u_t(x_t \mid z) = \frac{\dot{\sigma}_t}{\sigma_t}x_t + (\dot{\alpha}_t - \alpha_t \frac{\dot{\sigma}_t}{\sigma_t})z
    \label{eq:ConditionalVF}
\end{equation}
By construction, simulating an ODE with the conditional vector field $u_t(x\mid z)$ yields a trajectory that follows the Gaussian conditional probability path $p_t(x_t\mid z)$, thus the (Liouville \cite{ArnoldMMC}) continuity equation for the ODE described by $u_t(x_t\mid z)$ holds:
\newcommand{\dd}{\mathop{}\!\mathrm{d}}
\begin{multline}
    X_0 \sim \mathcal{N}\big(0, I_d\big), \quad \frac{\dd}{\dd t}X_t = u_t\big(X_t\mid z\big) \\ \implies X_t \sim p_t(\cdot\mid z) \quad \forall \quad t \in [0,1]\\
    \iff \partial_tp_t(x\mid z) = - \operatorname{div}_x\big(u_t(x_t \mid z)p_t(x_t \mid z)\big)
    \label{eq:condcontinuity}
\end{multline}
Corresponding \textbf{marginal probability path}, $p_t(x_t) = \mathbb{E}_{z \sim p_{\text{data}}}[p_{t}(x_{t}\mid z)]$ is induced by making $z \sim p_{\text{data}}$ random. The marginal probability path interpolates Gaussian noise $P_{0} \sim \mathcal{N}(0, I_{d})$ and $p_1 = p_{\text{data}}$. Marginalizing \eqref{eq:condcontinuity} over $z \sim p_{\text{data}}$ yields the marginal continuity equation\cite{Lipman2024FMGuide}:
\begin{multline}
    X_0 \sim \mathcal{N}\big(0, I_d\big), \quad \frac{\dd}{\dd t}X_t = u_t(X_t) \\ \implies X_t \sim p_t \quad \forall \quad t \in [0,1]\\
    \iff \partial_tp_t(x) = - \operatorname{div}_x\big(u_t(x_t)p_t(x_t)\big)
    \label{eq:marginalcontinuity}
\end{multline}
Flow Matching models learn to approximate the \textbf{marginal vector field} $u_{t}(x_t)$ using a deep neural network, Mamba backbone in particular for the scope of this paper:
\begin{multline}
u_t(x_t) = \int u_t(x_t \mid z)\, p_{1\mid t}(z \mid x_t)\,\mathrm{d}z, \\
p_{1|t}(z \mid x_t) = \frac{p_t(x_t \mid z)p_{\text{data}}(z)}{p_t(x_t)}
\label{MarginalVectorField}
\end{multline}
\paragraph{Conditional Flow Matching (CFM).}
Directly regressing the marginal field $u_t(x)$ via \eqref{MarginalVectorField} is intractable because it
requires the posterior $p_{1|t}(z\mid x_t) \propto p_t(x_t\mid z)p_{\text{data}}(z)$.
Instead, CFM supervises the network with the \emph{conditional} target $u_t(x_t\mid z)$, which
admits a closed form under Gaussian probability paths (Eq.~\ref{eq:ConditionalVF}).
We draw $t\!\sim\!\rho$, $z\!\sim\!p_{\text{data}}$, $\epsilon\!\sim\!\mathcal{N}(0,I)$ and set
$x_t=\alpha_t z + \sigma_t \epsilon$, then minimize
\begin{equation}
\label{eq:cfm-loss}
\mathcal{L}_{\text{CFM}}(\theta)
=\mathbb{E}_{t\sim\rho, z\sim p_{\text{data}}, \epsilon}
\left[\;\big\|u_\theta(x_t,t)-u_t(x_t\mid z)\big\|_2^2\;\right].
\end{equation}
Training on the conditional target sidesteps the intractable posterior in the marginal formula. With an $\ell_2$ loss, the best predictor at any $x_t$ is the conditional mean of $u_t(x_t\mid z)$ over $z \sim p_{\text{data}}$, which equals the marginal vector field $u_t(x_t)$.
i.e., CFM learns the \emph{marginal} vector field without ever evaluating the intractable posterior $p_{1|t}(z\mid x_t)$.
In practice we use $\rho(t)\!=\!\mathrm{Uniform}[0,1]$ and the GVP interpolant $(\alpha_t,\sigma_t)$ (Sec.~\ref{sec:experiments}).

Simulating an ODE with the marginal vector field from initial Gaussian noise leads to a trajectory whose marginals at $p_t$, such that $X_1 \sim p_{\text{data}}$, see Eq.\eqref{eq:marginalcontinuity}, returns a sample from the desired distribution. This sampling is called \textbf{ODE sampling} with a flow matching model, analogous in spirit to ODE-based samplers developed for diffusion models \cite{Lu2022DPMSolver}.

\subsection{Selective State Space Models: Mamba}
\paragraph{From structured SSMs to Mamba.}
A continuous-time linear state-space model (SSM) evolves
\begin{equation}
\dot{h}(t)=A\,h(t)+B\,x(t), \qquad y(t)=C\,h(t)+D\,x(t),
\label{eq:cts-ssm}
\end{equation}
where $h\in\mathbb{R}^{d_{\text{inner}}\times d_{\text{state}}}$, input $x\in\mathbb{R}^{d_{\text{inner}}}$, and output $y\in\mathbb{R}^{d_{\text{inner}}}$.
Classical \emph{structured} SSMs use fixed, specially parameterized $(A,B,C,D)$ and model long-range dependencies within a sequential signal via a causal convolution kernel in discrete time.
\paragraph{Zero-order hold (ZOH) discretization.}
With a (possibly learned) positive step $\Delta>0$ and zero-order hold assumption on the input over $[k\Delta,(k{+}1)\Delta)$, the exact discretization is
\begin{equation}
h_{k+1}=\bar A\,h_k+\bar B\,x_k,\qquad y_k=C\,h_k+D\,x_k,
\label{eq:zoh}
\end{equation}
where
\begin{equation}
\bar A \;=\; e^{\Delta A}, 
\qquad \bar B \;=\; \left(\int_0^{\Delta}\! e^{\tau A}\,d\tau\right) B.
\label{eq:zoh-AB}
\end{equation}

\paragraph{Selective (input-dependent) SSM.}
Mamba~\cite{GuDao2023Mamba} relaxes the time-invariance constraint in SSMs making the parameters B, C and step size $\Delta$ functions of input $x(t)$ at each time step t, thereby making the state-space dynamics content aware
\begin{equation}
B_k,C_k,\Delta_k \;\;=\;\; \Phi(x_k\,\theta),
\label{eq:selective-params}
\end{equation}
yielding a time-varying discrete recurrence that facilitates expressive causal aggregation of information across input signal 
\begin{equation}
h_{k+1}=\bar A_k\,h_k+\bar B_k\,x_k,\qquad y_k=C_k\,h_k+D_k\,x_k,
\label{eq:tv-recurrence}
\end{equation}
with $\bar A_k=e^{\Delta_k A_k}$ and $\bar B_k=(\int_0^{\Delta_k} e^{\tau A_k}\,d\tau) B_k$.

\subsubsection{Selective scan}
Let $u\in\mathbb{R}^{T\times d_{\text{inner}}}$ be a discrete time input signal and assuming the selective scan at step $t$ uses time variant SSM parameters $(\bar A_t = e^{\Delta_tA},\bar B_t,C_t)$ (Eq.~\ref{eq:selective-params}). The selective scan evaluates the causal recurrence
\begin{equation}
\label{eq:ssm-scan}
\begin{aligned}
h_{t+1} = \bar A_t h_t + \bar B_t\, u(t), \\
y(t)     = C_t h_t + D_t\, u(t), \\
h_0     = 0 \,.
\end{aligned}
\end{equation}

\paragraph{Unrolled recurrence.}
Expanding \eqref{eq:ssm-scan} yields
\begin{multline}
\label{eq:unrolled-h}
h_{t}=\Big(\!\prod_{j=0}^{t-1}\!\bar A_j\Big) h_0
\;+\; \sum_{i=0}^{t-1}\Big(\!\prod_{j=i+1}^{t-1}\!\bar A_j\Big)\bar B_i u(i)\\
h_0 = 0 \implies h_t
\;=\; \sum_{i=0}^{t-1}\Big(\!\prod_{j=i+1}^{t-1}\!\bar A_j\Big)\bar B_i u(i),   
\end{multline}

and therefore
\begin{equation}
\label{eq:unrolled-y}
y(t)
= C_t h_t + D u(t)
= D u(t) \;+\;C_t\sum_{i=0}^{t-1}
\Big(\!\prod_{j=i+1}^{t-1}\!\bar A_j\Big)\bar B_i u(i).
\end{equation}
where empty product is $I$ (identity). Equation \eqref{eq:unrolled-y} implies that the discrete time output signal $y(t)$ depends only on $\{x(0),\dots,x(t)\}$.

\paragraph{Hardware-aware evaluation.}
Rather than forming long products in \eqref{eq:unrolled-h}, Mamba uses a fused prefix (monoid) scan on pairs $(\bar A_k,\bar B_k x_k)$ composed as
$(\bar A',\bar B')\circ(\bar A,\bar B)=(\bar A'\bar A,\;\bar A'\bar B+\bar B')$,
yielding linear-time aggregation and high GPU efficiency.

\subsection{Motivation and Method}
It has been established that State Space Models are effective signal approximators \cite{WangXue2024UniversalSSM}. In particular, due to their time variant selective scan operation, Mamba SSMs have shown promising results for efficient long sequence modeling tasks such as, tokenization free byte-based language modeling \cite{Wang2024MambaByte}, modeling audio waveforms and DNA sequences \cite{GuDao2023Mamba}. 

\subsubsection{Selective Scan Operation: Causality}
Mamba aggregates information across input signal in an input dependent causal manner via the efficient selective scan operation, thus excelling at efficient long sequence modeling. Assume $u \in \mathbb{R}^{T \times d_{\text{inner}}}$ is the discrete time input signal and $y \in \mathbb{R}^{T \times d_{\text{inner}}}$ is the discrete time output of selective scan operation.
Since from equation (\ref{eq:unrolled-y}) it is evident that $y(t)$ depends only on $\{x(0),\dots,x(t)\}$,  we can formalize the conventional selective scan manifold (Fig.~\ref{fig:selectivescanmanifold}(a)), where state-space dynamics are always initialized with \emph{0} initial value condition and terminal state-space representation $h_T$ \footnote{We denote the terminal SSR as $h_T$; under zero-based indexing this is equivalently $h(T\!-\!1)$.} is discarded after driving $y(T-1)$, as follows:
\paragraph{Conventional selective-scan manifold:}
Under \eqref{eq:ssm-scan}–\eqref{eq:unrolled-y} with the fixed initial condition $h(0)=\mathbf{0}$, a Mamba block implements a causal map
\begin{multline}
\mathcal{M}:\; u \mapsto y,\qquad
y(t) = \mathcal{F}\big(u(0{:}t)\big),\quad t=0,\dots,T\!-\!1, \\ h_0 = 0
\label{eq:manifold-conventional}
\end{multline}

and discards the terminal state-space representation $h_T$ (otherwise denoted as $h(T-1)$) after producing $y(T\!-\!1)$ (Fig.~\ref{fig:selectivescanmanifold}(a)).
Further, from (\ref{eq:unrolled-y}) the Jacobian $J=\big[\partial y(i)/\ \partial u(j)\big]$

\[
\frac{\partial y(j)}{\partial u(i)}=
\begin{cases}
D, & j=i,\\[2pt]
C_j\!\left(\prod_{k=i+1}^{j-1}\bar A_k\right)\!\bar B_i, & i<j,\\[4pt]
\mathbf{0}, & i>j,
\end{cases}
\]
is strictly lower-triangular, no input $u(i)$ can contribute to output $y(j)$ with $i>j$.

This causal aggregation of information causes friction when adapting Mamba to non-sequential signals (such as images), when flattened in particular scan order and subjected to selective scan operation, motivating architectures with composite inductive biases that relax the strict causality of selective scan operation at scale throughout the Deep Neural Network (DNN).

Prior works instrument the same conventional selective scan manifold (\ref{eq:manifold-conventional}), combined with different scan order permutations. One approach (Zhu \etal, \cite{visionMamba}) proposes aggregation over the non-sequential input signal in different scan orders with different sets of SSM weights, followed by simple feature fusion. Another approach amortizes layerwise heterogeneous scan order causal aggregations across depth of the DNN with no per-block parameter increase, as in \emph{Zigma} \cite{ZigMaECCV}, where each selective scan manifold $u \mapsto y$, captures dependencies between tokens at varying degrees of spatial vicinity (specified by the particular scan orders at corresponding layers) across layers in the DNN.

\subsubsection{Generalizing the selective-scan manifold}
Given the fact that the selective-scan operation aggregates information across input signal in a strictly causal manner, evolving the latent state-space representation (SSR) as Eq.~(\ref{eq:tv-recurrence}), we hypothesize that the terminal SSR $h_T$ computed during conventional selective scan Eq.~(\ref{eq:manifold-conventional}) through a full causal pass over the non-sequential input signal, in specified scan order, potentially encodes a useful global summary and serves as a mild inductive cue for downstream state-space dynamics for subsequent blocks, despite the SSR being a severe compression of the non-sequential signal. We therefore generalize the conventional selective-scan manifold (\ref{eq:manifold-conventional}) with a Differentiable \textbf{R}ecurrent State \textbf{C}hain to \emph{Arcee} selective-scan manifold (see Fig.~\ref{fig:selectivescanmanifold}(b)) that accepts initial value condition for internal state-space evolution dynamics and exposes it's terminal SSR $h_T$ for upstream computation.

\paragraph{Arcee selective-scan manifold:}
We extend the conventional map \eqref{eq:manifold-conventional} to a two-port block that \emph{accepts} an initial state and \emph{returns} its terminal state:

\begin{multline}
\label{eq:manifold-arcee}
\mathcal{M}^{(\ell)}:\; \big(u^{(\ell)}(\cdot),\,h^{(\ell)}(0)\big)\;\longmapsto\; \big(y^{(\ell)}(\cdot),\,h^{(\ell)}(T)\big),
\\ y^{(\ell)}(t) = \mathcal{F}\big(u^{(\ell)}(0{:}t)\big), \quad \forall~ t\in[0, T-1]
\end{multline}
where the intra-block dynamics remain the causal selective scan of \eqref{eq:ssm-scan} (with ZOH factors $\bar A_t=e^{\Delta_t A}$, $\bar B_t=\big(\!\int_0^{\Delta_t}\! e^{\tau A} d\tau\big)B$).

\paragraph{Differentiable boundary map and cross block chaining:}
We connect Mamba blocks by a differentiable boundary map that seeds the next block’s initial condition with the previous block’s terminal SSR:
\begin{multline}
\label{eq:boundary-map}
h^{(\ell)}(0) \;=\; \mathcal{T}^{(\ell)}\!\big(h^{(\ell-1)}(T-1)\big),\\ \mathcal{T}^{(\ell)}=\mathrm{Identity}\ \text{by default}.    
\end{multline}

Even with such differentiable recurrent state chain, each block is still strictly causal internally and the $u^{(\ell)} \mapsto y^{(\ell)}$ mapping satisfies the lower-triangular Jacobian of the causal scan:
\[
\frac{\partial y^{(\ell)}(j)}{\partial u^{(\ell)}(i)}=
\begin{cases}
D^{(\ell)}, & j=i,\\[2pt]
C^{(\ell)}_j\!\Big(\prod_{k=i+1}^{j-1}\bar A^{(\ell)}_{k}\Big)\!\bar B^{(\ell)}_{i}, & i<j,\\[4pt]
\mathbf{0}, & i>j,
\end{cases}
\] Where empty product is $I$ (identity).

Composing a DNN using $L$ Mamba blocks with \emph{Arcee} modification yields cross-depth system as follows:
\begin{equation}
\label{eq:chain}
\begin{split}
(u^{(0)},\dots,u^{(L-1)}) \longmapsto (y^{(0)},\dots,y^{(L-1)}),\\
h^{(\ell)}(0)=\mathcal{T}^{(\ell)}\!\big(h^{(\ell-1)}(T-1)\big),\ \forall\,\ell\in\{1,\dots,L-1\},\\
h^{(0)}(0)=0 \quad \text{for block 0}.
\end{split}
\end{equation}

\paragraph{Cross-block Jacobian (Arcee).}
Because $h^{(\ell)}(0)=\mathcal{T}^{(\ell)}(h^{(\ell-1)}_T)$, outputs of block $\ell$ depend on inputs of block $\ell{-}1$ only via the terminal handoff. Let
$\;J_T^{(\ell)} \!=\! \partial h^{(\ell)}(0)/\partial h^{(\ell-1)}_T\;$ be the boundary Jacobian. Then for $j\in[0,T\!-\!1]$ and $i\in[0,T\!-\!1]$,
\begin{multline}
\label{eq:cross-block-jac}
\frac{\partial\, y^{(\ell)}(j)}{\partial\, u^{(\ell-1)}(i)}
\;=\;
\underbrace{C^{(\ell)}_j \!\!\left(\prod_{k=0}^{\,j-1}\!\bar A^{(\ell)}_{k}\right)}_{\text{downstream readout through block }\ell}\\
\;\; J_T^{(\ell)} \;\;
\underbrace{\left(\prod_{k=i+1}^{\,T-1}\!\bar A^{(\ell-1)}_{k}\right)\bar B^{(\ell-1)}_{i}}_{\text{upstream contribution to }h_T^{(\ell-1)}}.    
\end{multline}

(Empty products equal the identity $I$.) Equation~\eqref{eq:cross-block-jac} consists of three parts:
(i) an \emph{upstream causal accumulator} that builds the terminal state $h_T^{(\ell-1)}$ inside block $\ell\!-\!1$;
(ii) a \emph{boundary map} $J_T^{(\ell)}=\partial h^{(\ell)}(0)/\partial h_T^{(\ell-1)}$ that hands off this terminal state to the next block (we default to $\mathcal{T}^{(\ell)}{=}\mathrm{Identity}$);
and (iii) a \emph{downstream causal propagator} inside block $\ell$ that maps $h_0^{(\ell)}$ to outputs $y^{(\ell)}$.
Therefore, each block stays strictly causal \emph{inside} itself; only the compact terminal state is passed across blocks.

\paragraph{Network-level Jacobian and implicit limitations.}
If we stack blocks by depth, the overall Jacobian is \emph{block–lower–triangular}:
outputs of block $\ell$ never depend on inputs of any future block $m{>}\ell$.
Any cross–block influence from block $m$ to $\ell$ must pass through the
terminal–state handoff $h_T\!\in\!\mathbb{R}^{d_{\text{inner}}\times d_{\text{state}}}$,
which acts as a low–dimensional bottleneck.
Consequently, the off–diagonal Jacobian blocks are low–rank; a safe, concise bound is
\begin{equation}
\label{eq:ssrbottleneck}
\operatorname{rank}\!\left(\frac{\partial y^{(\ell)}}{\partial u^{(m)}}\right)
\;\le\; d_{\text{inner}} \cdot d_{\text{state}}
\qquad\text{for } m<\ell.
\end{equation}

i.e., cross–block effects are compressed by the SSR and cannot carry full per-token detail.

Equivalently, the implication is that each selective scan manifold throughout the DNN, starts with the terminal SSR $h_T$ from previous block as a directional prior reflecting a full causal pass over input signal (in specified scan order). Although $h_T$ is a severe compression, we hypothesize it still encodes useful global summary. As a consequence, the differentiable recurrent state chain effectively acts as a architecture agnostic composite inductive bias across depth for Mamba based DNNs, in the sense that it can be plugged into selective-scan operation for any Mamba based DNN. Although the mamba blocks in isolation still remain causal, \emph{Arcee} alleviates the strict causality of the selective-scan operation across the depth of the DNN, thereby increasing performance on non-sequential modalities as evident from our experiments discussed in results.

\subsubsection{Implementation details (Arcee modifications)}
We modify both the selective scan forward/backward fused kernels (introduced in \cite{GuDao2023Mamba}) to align with \emph{Arcee selective-scan manifold} (\ref{eq:manifold-arcee})
\paragraph{Forward (one read, one write):}
\[
h^{(\ell)}(0)\leftarrow \mathcal{T}^{(\ell)}\!\big(h^{(\ell-1)}(T-1)\big),\qquad
\mathcal{T}^{(\ell)}=\mathrm{Id}\ \text{by default}.
\]
For the fused forward CUDA kernel, this adds a single read of $h^{(\ell-1)}(T-1)$ and a single write to $h^{(\ell)}(0)$.

\paragraph{Backward (seeded terminal adjoint):}
Let $g_{y^{(\ell)}(t)} = \partial\mathcal{L}/\partial y^{(\ell)}(t)$ and
$g_{h^{(\ell)}(t)} = \partial\mathcal{L}/\partial h^{(\ell)}(t)$. 
The only change vs. vanilla is the \emph{initial seed} for the terminal adjoint at block $\ell-1$ due to the boundary handoff into block $\ell$:
\begin{multline}
\label{eq:seed}
g_{h^{(\ell-1)}(T-1)} \;\mathrel{+}=\
\underbrace{\frac{\partial \mathcal{L}}{\partial y^{(\ell-1)}}\!
\frac{\partial y^{(\ell-1)}}{\partial h^{(\ell-1)}(T-1)}}_{\text{local (within block $\ell-1$)}}
\;+\; \\
\underbrace{ \big(J_{\mathcal{T}}^{(\ell)}\big)^{\!\top}\, g_{h^{(\ell)}(0)} }_{\text{boundary from next block}},    
\end{multline}

where $J_{\mathcal{T}}^{(\ell)}=\partial h^{(\ell)}(0)/\partial h^{(\ell-1)}(T-1)$.  
For the default identity mapping, $J_{\mathcal{T}}^{(\ell)}=I$ and the boundary term reduces to $g_{h^{(\ell)}(0)}$.
Given the terminal seed \eqref{eq:seed}, the per-token adjoint recurrences inside block $\ell$ are identical to conventional selective-scan:
\begin{equation}
\label{eq:bwd-core}
\begin{aligned}
g_{h^{(\ell)}(t)} &\mathrel{+}= (C_t^{(\ell)})^\top\, g_{y^{(\ell)}(t)}
                   \;+\; (\bar A_t^{(\ell)})^\top\, g_{h^{(\ell)}(t+1)},\\
g_{u^{(\ell)}(t)} &\mathrel{+}= (D_t^{(\ell)})^\top\, g_{y^{(\ell)}(t)}
                   \;+\; (\bar B_t^{(\ell)})^\top\, g_{h^{(\ell)}(t+1)}.
\end{aligned}
\end{equation}
Parameter gradients accumulate as usual via chain rule through the selective heads:
$(\bar A_t^{(\ell)},\, \bar B_t^{(\ell)},\, C_t^{(\ell)},\, D_t^{(\ell)},\, \Delta_t^{(\ell)})
= \Phi\!\big(u(t)^{(\ell)};\,\theta\big).$

\paragraph{Cost and memory.}
Arcee adds $O(d_{\text{state}})$ work per block from the boundary read/write and the terminal seed in \eqref{eq:seed}. The scan FLOPs and memory traffic remain $O(T\cdot d_{\text{state}})$, which is identical to the vanilla fused selective scan. No additional activations are required beyond storing $h^{(\ell)}(T-1)$.

\paragraph{Stability note.}
With Hurwitz $A$ and $0<\Delta_{\min}\le \Delta_t\le\Delta_{\max}$, we have $\rho(e^{\Delta_t A})<1$ and the product $\prod_t e^{\Delta_t A}$ remains bounded, thus the SSM is stable for any $h_0$.

\section{Experiments}
\label{sec:experiments}

\subsection{Setup}
\label{sec:setup}
We test the hypothesis that the terminal state–space representation $h_T$ from a causal Mamba pass over a flattened, non-sequential signal acts as a mild \emph{directional prior}: a compact global summary that, when reused across depth as initial value condition for state-space dynamics in subsequent block, improves downstream selective-scan dynamics yielding a composite inductive bias at scale that reduces friction when adapting Mamba blocks to non-sequential signals despite their strict causality in isolation.

\textbf{Framework.}
We evaluate \emph{Arcee} within the \emph{Flow Matching} framework for unconditional image generation. We parameterize the \emph{marginal vector field} $u_t(\cdot)$ (see Eq.~(\ref{MarginalVectorField})) with Mamba-based DNNs that process non-sequential signals (images in this case) as flattened tokens via selective scans.

Concretely, we compare Mamba-based DNNs that use the conventional selective scan (see Fig.~\ref{fig:selectivescanmanifold}(a), (\ref{eq:manifold-conventional})) against the same DNNs where mamba blocks are augmented with the \emph{Arcee} selective scan (a recurrent state chain that reuses terminal SSR as the initial condition for SSM dynamics in subsequent block through a differentiable boundary map across depth; see Fig.~\ref{fig:selectivescanmanifold}(b), (\ref{eq:manifold-arcee})). To isolate the hypothesis under a fixed compute budget, we use a single dataset (CelebA-HQ $256{\times}256$) and omit unrelated backbones (e.g., Transformers/UNets \cite{Ronneberger2015UNet, Dosovitskiy2021ViT, PeeblesDiT}).

\textbf{Deep Neural Network (DNN) Backbones.}
We integrate \emph{Arcee} as a \emph{boundary hook} that seeds each
block’s initial state $h^{(\ell)}(0)$ with the previous block’s terminal
SSR $h^{(\ell-1)}_T$ (Fig.~\ref{fig:selectivescanmanifold}(b),
see Eq.~\ref{eq:manifold-arcee}), leaving the selective-scan body
unchanged. This makes \emph{Arcee} orthogonal to backbone specifics;
we evaluate its effect on two Mamba-based DNN backbones:
(i) \textbf{Zigzag-Mamba (Zigma)}~\cite{ZigMaECCV}, which amortizes
heterogeneous scan-order permutations across depth so different Mamba
blocks model relationships between tokens at varying spatial vicinities;
(ii) \textbf{Vision Mamba}~\cite{visionMamba}, which employs
order-specific SSM weights followed by simple feature fusion.

We use Zigzag-Mamba (Zigma) and Vision Mamba in their original forms
as published in \cite{ZigMaECCV,visionMamba}, without any architectural
modification. Arcee is implemented purely as an extension of the
conventional selective-scan manifold via a recurrent terminal-SSR
chain and a differentiable boundary map across depth; no architectural
diagrams change, only the $\big(h^{(\ell-1)}_T, h^{(\ell)}(0)\big)$ boundary
is enabled (see Fig.~\ref{fig:selectivescanmanifold}).

\textbf{Training details (Flow Matching).}
For the interpolation schedulers in Eq.~(\ref{eq:ConditionalProbPath})
we use the generalized VP (GVP) interpolant from SiT~\cite{Ma2024SiT}
with $\alpha_t = \cos\!\big(\tfrac{\pi}{2} t\big)$ and
$\sigma_t = \sin\!\big(\tfrac{\pi}{2} t\big)$, sampling
$t \sim \mathrm{Uniform}[0,1]$.
Targets follow the conditional Flow Matching objective
(Eq.~\ref{eq:cfm-loss}).
Unless noted otherwise, we train with AdamW (no weight decay),
a constant learning rate of $3{\times}10^{-4}$, global batch size
$192$, image resolution $256^2$, and $50{,}000$ optimization steps; we
enable RMSNorm, fused add–norm, and learnable positional embeddings,
and set $d_{\text{state}}{=}256$.
Budgets (steps, batch, sampler settings) are matched across each
baseline and its +Arcee counterpart; Arcee uses
$\mathcal{T} = \mathrm{Id}$ and adds no extra parameters.
All experiments are conducted in the latent space of a pre-trained
variational autoencoder (VAE)~\cite{kingma2014autoencoding,sdvaeftema}
with compression factor 8. We maintain an exponential moving average (EMA) of parameters with decay rate $\beta = 0.9999$ and report results using the EMA weights.

\textbf{DNN configurations.}
For each DNN backbone we report depth $L$, embedding $d_{\text{model}}$, and parameter count.
The state size is fixed to $d_{\text{state}=256}$ for all experiments. The +Arcee variants keep all hyperparameters fixed; only the $(h_0,h_T)$ boundary is enabled (no extra params). To compensate for parameter overhead, per additional scan-order, due to order specific SSM weights, we reduce depth of the vision mamba backbone \cite{visionMamba} to match parameter count with Zigma backbones.
\begin{table}[t]
\centering
\caption{Backbone specs (identical for baseline vs.\ +Arcee).}
\label{tab:backbones}
\begin{tabular}{lccc}
\toprule
Family & depth & $d_{\text{model}}$ & Params (M) \\
\midrule
Zigma-$k$ ($k\in\{1,2,4,8\}$) & 24 & 768 &  161.8 \\
Vision Mamba                  & 20 & 768 &  161.9  \\
\bottomrule
\end{tabular}
\end{table}

\textbf{Evaluation.}
We report \textbf{FID}$\downarrow$ (CleanFID) on $50$K samples and
\textbf{KID}$\downarrow$, both computed with Inception-V3 features
\cite{parmar2022cleanfid,binkowski2018demystifying,szegedy2016rethinking}.
Sampling uses the same ODE solver (Dormand--Prince, \texttt{dopri5})
and a fixed number of function evaluations (NFE = 50) across all
models~\cite{dormand1980prince}.
All models are trained for exactly $50{,}000$ steps, and we evaluate
the EMA checkpoint at this fixed step for every baseline and +Arcee
variant.

\subsection{Main results}
\label{sec:main-results}

\begin{table}[t]
\centering
\caption{CelebA-HQ (256$^2$) Flow Matching. FID/KID: lower is better.
All models trained for 50K steps with matched budgets; +Arcee variants
add no parameters.}
\label{tab:main}
\begin{tabular}{lcc}
\toprule
Model & FID$\downarrow$ & KID $\times 10^3 \downarrow$ \\
\midrule
Zigma-1 (baseline)         & 82.81 &   88.69 \\
\;\;+\;Arcee (ours)      & \textbf{15.33} & \textbf{10.59}   \\
\midrule
Zigma-4 (baseline)         & 11.27 & 7.70 \\
\;\;+\;Arcee (ours)      & \textbf{10.86} & \textbf{7.45} \\
\midrule
Vision Mamba (baseline)    & 14.08 & 9.76 \\
\;\;+\;Arcee (ours)      & \textbf{13.47} & \textbf{9.36} \\
\bottomrule
\end{tabular}
\end{table}

\noindent\textbf{Observation.}
On the naive single scan-order Zigma baseline, which is effectively strictly causal even across depth given the fact that all blocks employ the same scan-order to evaluate inherently selective-scan operation (\ref{eq:manifold-conventional}) over input signal, Arcee yields a large
improvement: FID drops from 82.81 to 15.33 ($5.4\times$ lower) and KID
from 88.69 to 10.59 as shown in Table~\ref{tab:main}, without changing the architecture or parameter
count.
For Zigma-4 (each block evaluates selective-scan over input in one of 4 different scan orders across depth; refer \cite{ZigMaECCV} for details), Arcee still provides a consistent gain (11.27 $\rightarrow$
10.86 FID; 7.70 $\rightarrow$ 7.45 KID), indicating that cross-block
SSR reuse remains beneficial even when some scan-order diversity is
present.
Fusion of selective-scan evolution over the input in multiple scan
orders, each with its own SSM weights, already induces a strong
inductive bias in Vision Mamba, making the causal selective scan much
more amenable to non-sequential signals. As a result, Arcee yields a
smaller but still consistent improvement (14.08 $\rightarrow$ 13.47
FID; 9.76 $\rightarrow$ 9.36 KID) on top of this architecture, suggesting that propagating the causal directional prior ($h^{(\ell)}(0) \leftarrow h^{(\ell - 1)}_T$, refer Fig.~\ref{fig:selectivescanmanifold}(b), Eq.~\ref{eq:manifold-arcee}) can help across different Mamba-based DNNs.

\subsection{Baselines and ablations}
\label{sec:ablations}

\noindent\textbf{Baselines.}
We consider two families of Mamba-based DNN backbones:
(1) \textit{Zigma-$k$}~\cite{ZigMaECCV}, a Zigzag-Mamba variant with
$k$ scan orders amortized across depth; and
(2) \textit{Vision Mamba}~\cite{visionMamba}, which employs fusion of selective-scan evolution over input signal in multiple scan orders each with different sets of weights, thereby incurring parameter overhead for each additional scan-order.
For each backbone, we train the baseline model and its +Arcee
counterpart under identical budgets (Sec.~\ref{sec:setup}).

\noindent\textbf{Scan-order diversity (Zigma).}
To study the interaction between Arcee and scan-order diversity, we
vary the number of Zigma scan orders $k \in \{1,2,4,8\}$ and train
each configuration with and without Arcee. 

\begin{table}[t]
\centering
\caption{Effect of scan-order diversity on Zigma with and without
Arcee (CelebA-HQ 256$^2$). All models are trained for 50K steps with
matched budgets.}
\label{tab:zigma-scan-ablation}
\begin{tabular}{lccc}
\toprule
Model        & Orders $k$ & FID$\downarrow$ & KID $\times 10^3 \downarrow$ \\
\midrule
Zigma-1           & 1 & 82.81 & 88.69 \\
\;\;+\;Arcee     & 1 & \textbf{15.33} & \textbf{10.59} \\
\midrule
Zigma-2           & 2 & \textbf{15.64} & \textbf{11.22} \\
\;\;+\;Arcee     & 2 & 15.68 & 11.40 \\
\midrule
Zigma-4           & 4 & 11.27 & 7.70 \\
\;\;+\;Arcee     & 4 & \textbf{10.86}  & \textbf{7.45} \\
\midrule
Zigma-8           & 8 & \textbf{11.19} & \textbf{7.77} \\
\;\;+\;Arcee     & 8 & 11.75 & 8.17 \\
\bottomrule
\end{tabular}
\end{table}

We observe three regimes (see Table~\ref{tab:zigma-scan-ablation}). First, going from $k{=}1$ to $k{=}4$ improves
the baseline, but pushing to $k{=}8$ already hurts performance:
Zigma-8 is slightly worse than Zigma-4 under the same training budget,
indicating diminishing and eventually negative returns from adding more
scan orders to the baseline at fixed compute. Second, Arcee provides by far its largest
gain in the low-diversity setting ($k{=}1$), is effectively neutral
around $k{=}2$ (with small differences likely attributable to
optimization noise), and gives a modest but consistent improvement at
$k{=}4$. Third, in the heavily diversified setting $k{=}8$, Arcee-8
underperforms Zigma-8, suggesting that given the backbone’s scan-order
bias is already over-saturated and suboptimal, the additional
directional prior cannot compensate and may mildly interfere with it.

Overall, the best-performing configuration in this family is
Zigma-4+Arcee. This supports the view that Arcee is most useful in
low- to moderate-diversity regimes, where cross-block SSR reuse
complements existing architectural inductive biases by providing
additional directional context across depth, while very aggressive
scan-order diversification is evidently a poor design choice under our
setup.

\noindent\textbf{Cross-architecture generality (Vision Mamba).}
As shown in Tab.~\ref{tab:main}, Arcee also yields a smaller but
consistent improvement on Vision Mamba, despite its strong built-in
scan-order inductive bias via order-specific SSM weights. Together with
the Zigma results above, this suggests that the proposed directional
prior can be instrumented as a lightweight, plug-in boundary hook across
different Mamba-based backbones, with the largest benefits appearing
when global mixing is present but not already over-saturated by very
aggressive scan-order diversification.

\begin{figure}[t]
  \centering
  \includegraphics[width=\linewidth]{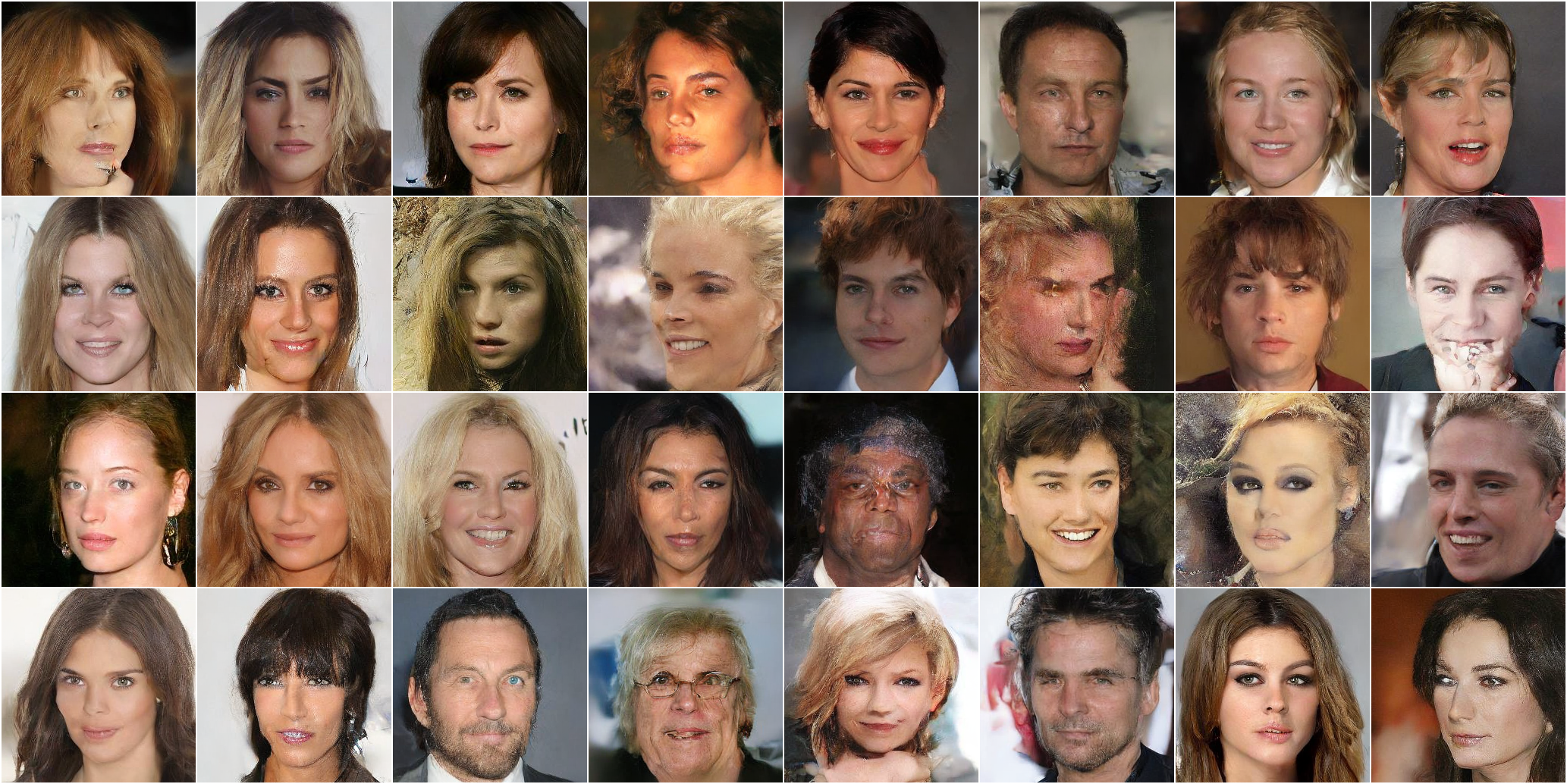}%
  \caption{Zigma-1 + Arcee Qualitative CelebA-HQ (256$^2$) samples.
  Arcee enables sharper, more coherent faces under the same sampling budget.}
  \label{fig:rcqual}
\end{figure}

\begin{figure}[t]
  \centering
  \includegraphics[width=\linewidth]{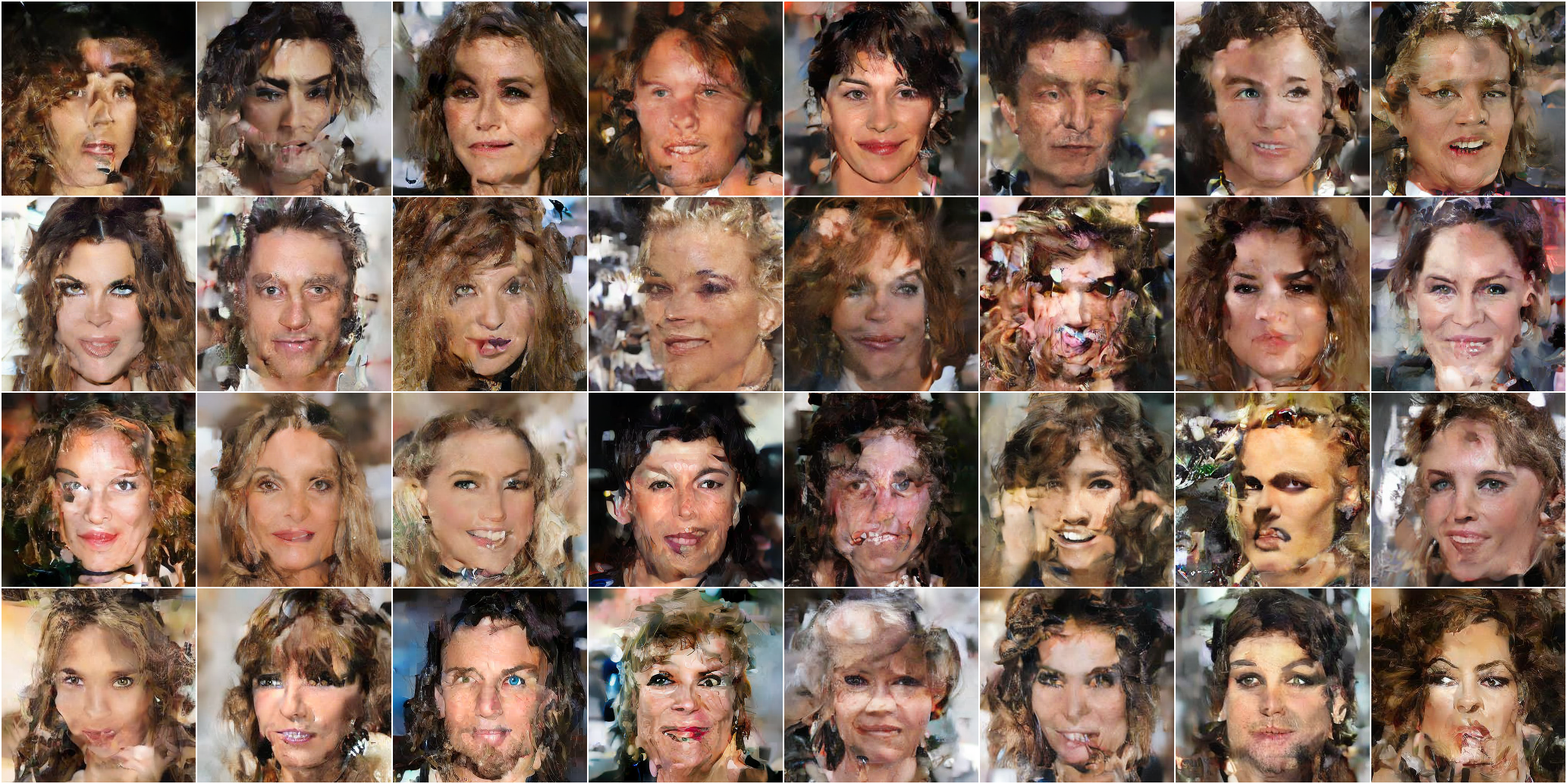}%
  \caption{Zigma-1 baseline Qualitative CelebA-HQ (256$^2$) samples.}
  \label{fig:zqual}
\end{figure}

\section{Conclusion}
We developed \emph{Arcee}, a structure-informed recurrent state chain
for Mamba-based DNNs that reuses the terminal state–space
representation $h_T$ as a cross-block directional prior. Leveraging
the underlying state–space dynamics, the design operates purely as a
boundary hook on the selective-scan manifold, leaving the internal
Mamba block, parameter count, and computational complexity unchanged.
By propagating $h_T$ across depth through a differentiable boundary
map, Arcee induces a mild yet global inductive bias that helps adapt
strictly causal selective scans to non-sequential signals.

Unlike prior ``Mamba-for-vision'' variants that rely primarily on scan
order permutations or order-specific SSM weights, the proposed Arcee
design is informed by the state–space structure itself: it explicitly
reuses the terminal SSR as a compact global summary across blocks,
while preserving per-block causality. Within the Flow Matching
framework on CelebA-HQ (256$^2$), Arcee yields an $81.5\%$ reduction
in FID (82.81 $\rightarrow$ 15.33) and an $88.1\%$ reduction in KID
(88.69 $\rightarrow$ 10.59) on the naive single-order Zigma baseline,
and provides consistent improvements for Zigma-4 ($\approx 3.6\%$ FID
and $3.2\%$ KID gains) and Vision Mamba ($\approx 4.3\%$ FID and
$4.1\%$ KID gains), all without adding parameters. Ablations over
scan-order diversity further reveal that Arcee is most beneficial in
low- to moderate-diversity regimes, while overly aggressive
scan-ordering can itself become suboptimal under fixed budgets.

\textbf{Future work.}
Arcee can be viewed as a generalization of the conventional
selective-scan manifold to a two-port interface
$(u, h^{(\ell)}(0)) \mapsto (u, h^{(\ell)}_T)$: it supports nonzero,
potentially learned initial value conditions for the state–space
dynamics inside each selective scan, while exposing the terminal SSR
$h_T$ through a differentiable boundary map for upstream computation.
This perspective suggests several directions, including conditioning
selective-scan evolution in Mamba blocks on cross-modal priors (e.g.,
text, audio, or high-level semantic signals) via learned initial
states, extending Arcee to conditional generative modeling and other
modalities such as video and audio, and developing a sharper
theoretical characterization of cross-block recurrence in
continuous-time state–space models.

{
    \small
    \bibliographystyle{ieeenat_fullname}
    \bibliography{main}
}


\end{document}